\documentclass{article}
\usepackage{spconf,amsmath,graphicx}
\usepackage{xcolor}
\usepackage{multirow}
\usepackage{booktabs}
\usepackage{subcaption}

\usepackage{amsmath}
\usepackage{amsfonts}
\usepackage{graphicx}
\usepackage{xcolor}
\usepackage{soul}

\usepackage{tikz}
\usepackage{pgfplots}
\usepackage{pgfplotstable}
\usepackage{filecontents}

\usepackage{algorithm}       %
\usepackage{algpseudocode}   %
\algdef{SE}[SUBALG]{Indent}{EndIndent}{}{\algorithmicend\ }%
\algtext*{Indent}
\algtext*{EndIndent}
\usepackage{mathtools}

\title{A simple HMM with self-supervised representations \\ for phone segmentation}
\name{Gene-Ping Yang, Hao Tang}
\address{Centre for Speech Technology Research, University of Edinburgh}

\begin{document}
\maketitle
\begin{abstract}
Despite the recent advance in self-supervised representations, unsupervised phonetic segmentation remains challenging.
Most approaches focus on improving phonetic representations with self-supervised learning, with the hope that the improvement can transfer to phonetic segmentation.
In this paper, contrary to recent approaches, we show that peak detection on Mel spectrograms is a strong baseline, better than many self-supervised approaches.
Based on this finding, we propose a simple hidden Markov model that uses self-supervised representations and features at the boundaries for phone segmentation.
Our results demonstrate consistent improvements over previous approaches, with a generalized formulation allowing versatile design adaptations.
\end{abstract}

\section{Introduction}
\label{sec:intro}

Unsupervised phone segmentation is typically the first step to understanding speech from an unknown language.
Phone segmentation and phonetic unit discovery should in principle mutually benefit each other---a better phone segmentation leads to phonetic units that vary less across instances, and a set of phonetic units that represent segments better leads to more consistent phone segmentation.
Based on this intuition, a model for unsupervised phone segmentation should include both the modeling of the content in the segments and the modeling at the boundaries.

Recent research in unsupervised phone segmentation has mostly rely on self-supervised models, particularly those with contrastive learning  \cite{kreuk20_interspeech,chorowski21b_interspeech,bhati2022unsupervised,Cuervo2022contrastive}.
These approaches typically involve learning to contrast two contiguous frames, followed by a peak detection algorithm to identify phone boundaries from learned features.
For these approaches to work well, the main assumption is the existence of sharp boundaries.
However, given that the representations are contextualized \cite{Liu2024predictive}, the difference for any two contiguous hidden vectors is less likely to be sharp; hence the hypothesis noted in \cite{bhati2022unsupervised} that there is a trade-off between phone classification and phone segmentation performance.
In other words, contextualized representations are great at modeling the content of segments \cite{yang2022autoregressive}, but perhaps bad at modeling sharp boundaries.

Another approach to phone segmentation is based on clustering.
A recent example is duration-penalized dynamic programming (DPDP) \cite{kamper21_interspeech,kamper2022word}, which uses self-supervised features and a predefined set of code vectors. 
DPDP incorporates a duration penalty to encourage longer segments, with code vectors either jointly trained with self-supervised models or derived from k-means. 
This approach again relies on the modeling of content in segments and lacks modeling of boundaries.
However, given that this approach works sufficiently well, modeling the content of segments with a frame-based approach, particularly when using self-supervised representations, can go a long way.

In this paper, we first show that phone boundaries are best modeled by Mel spectrograms.
Peak detection on Mel spectrograms alone can outperform peak detection on many other self-supervised representations.
Though somewhat surprising, peak detection on spectrograms for unsupervised phone segmentation dates back to \cite{Glass1988multi} and has been a strong baseline for several decades.   
For the modeling of content in segments, we adopt a similar approach to DPDP, training an HMM on top of self-supervised representations \cite{yehlearning2023}.
In fact, DPDP can be seen as a special case of running Viterbi on an HMM \cite{Juang1990segmental}.
The benefits of using an HMM are two folds.
First, we can integrate the modeling of boundaries into the transition probabilities of the HMM.
Second, in contrast to DPDP which runs an offline k-means independent of the Viterbi algorithm, our HMM can be trained jointly alongside other constraints (such as limiting the number of segments) and the modeling of the boundaries.

We evaluate our proposed HMMs on TIMIT \cite{garofolo1993timit} and Buckeye \cite{pitt2007buckeye} for unsupervised phone segmentation, using self-supervised features extracted from pre-trained HuBERT and wav2vec 2.0 models. 
Our HMMs consistently outperform peak detection and DPDP on self-supervised representations, highlighting the importance of jointly optimizing the centroids (mean vectors in the emission probability) with the segmentation process. 
Additionally, by incorporating boundary features from Mel spectrograms, we achieve performance on par with or better than other approaches that require training neural networks of several layers, e.g., \cite{Strgar2023phoneme}.
Our approach has the advantage of being simple and fast.

\section{Boundary Features in Mel Spectrogram}

\begin{figure}[t!]
    \centering
    \includegraphics[width=\linewidth]{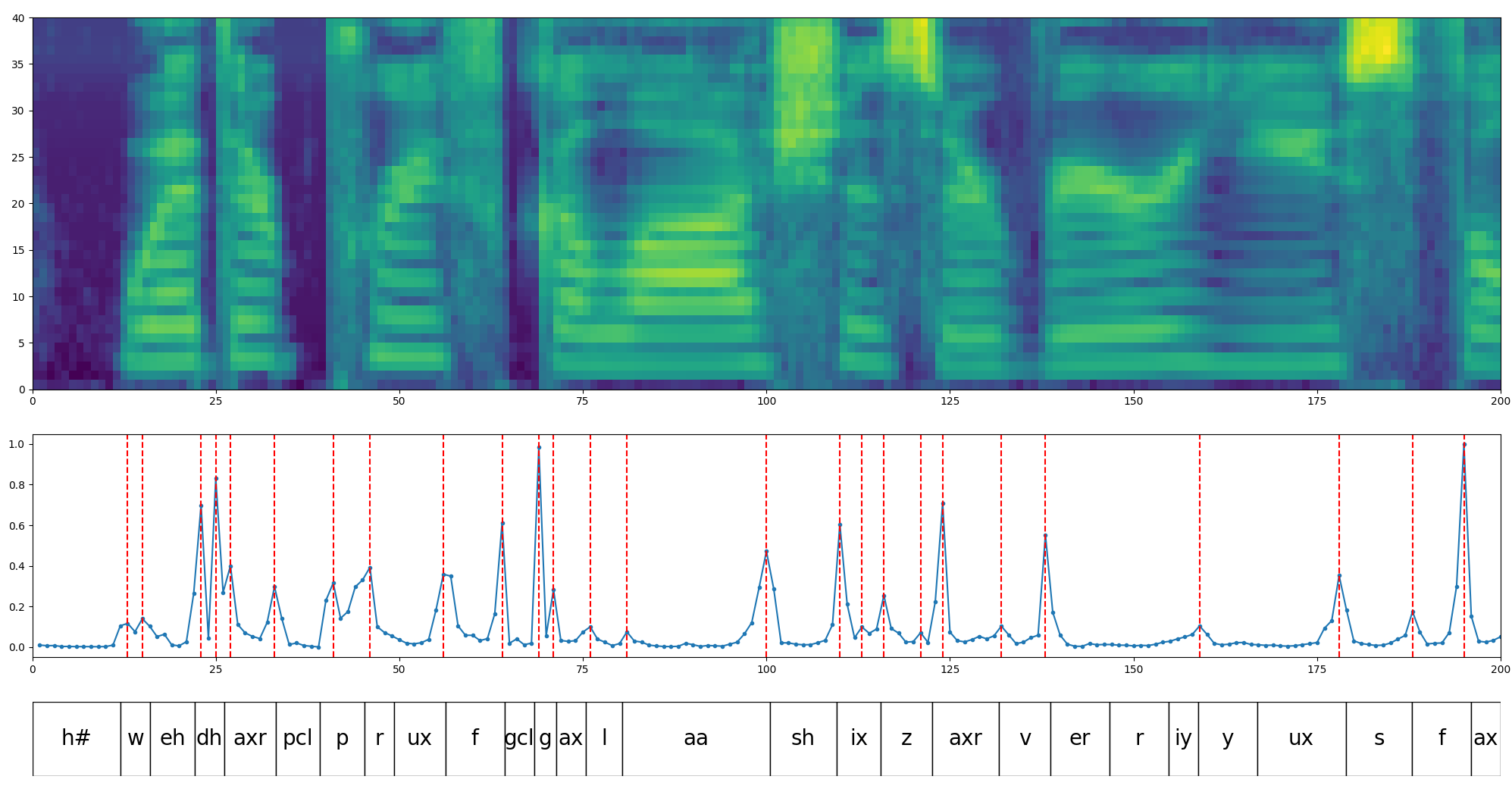}
    \caption{Peak detection using Mel spectrogram on the sample utterance fadg0\_sx289 from TIMIT. From top to bottom: Mel spectrogram, spectral variations, and ground truth phone segments.}
    \label{fig:logmel-peak}
    \vspace{-1em}
\end{figure}

A simple yet often overlooked method for unsupervised phone segmentation involves applying the spectral variation function (SVF) to Mel spectrograms or cepstral features.  
These features have been found highly correlated with phone boundaries \cite{Glass1988multi, flammia92_icslp, Mitchell_1995_using}.
These automatically discovered acoustic segments have been widely used to enhance HMM models in supervised phone recognition \cite{Mitchell_1995_using} and for unsupervised phone segmentation \cite{Wilpon1987investigation,Svendsen1987on,falavigna1990dtw,brugnara1993automatic,sharma96_icslp,rasanen2011blind,Hoang2011unsupervised,hoang2015blind,stan2016blind}.

One simple form of SVF uses the normalized spectral dot product (cosine distance)
\begin{equation}
    d_t = - \dfrac{x_{t-1}^\top x_{t}}{\|x_{t-1}\| \| x_{t} \|},
\end{equation}
\begin{equation}
    \tilde{d_t} = (d_t - d_{min}) / (d_{max} - d_{min}),
\end{equation}
where $x_t$ is the features at $t$, and $d_t$ measures the discrepancy between two contiguous frames, with a higher value indicating an abrupt acoustic event.
A peak detection algorithm is then perform on the normalized $\tilde{d_t}$ using topographical prominence\footnote{\scriptsize Peak detection is often implemented with \texttt{scipy.signal.find\_peaks}.}, and a threshold is used to identity the peaks with high prominence.
An illustration of the detected peaks is shown in Fig. \ref{fig:logmel-peak}.
Mel spectrograms possess a desirable property for peak detection algorithms---the spectral variations exhibit significant peaks while maintaining minimal variation within a phone segment.
Nevertheless, correctly identify phone boundaries between phones with smooth transitions, such as between a semi-vowel and a vowel, can be challenging.

\vspace{-0.8em}
\section{Applying HMMs to unsupervised phone segmentation}
\vspace{-0.5em}

Given the strong performance of peak detection on Mel spectrograms, in this section, we introduce an HMM to incorporate boundary features, self-supervised representations, and segmental constraints for unsupervised phone segmentation.
Many previous segmentation algorithms can be seen as HMMs.
Examples include Duration-Penalized Dynamic Programming (DPDP) \cite{kamper2022word, kamper21_interspeech} and Level Building Dynamic Programming (LBDP) \cite{sharma96_icslp, Myers1981lbdtw}.
In DPDP, the segmentation process is performed by minimizing the frame-wise distance between speech features to the closest VQ code vector while incorporating a duration penalty.
In Level Building Dynamic Programming \cite{sharma96_icslp, Myers1981lbdtw}, a constraint is set on the number of allowed segments.

\vspace{-0.8em}
\subsection{HMM Formulation}

We begin by formulating an HMM with segmental constraints imposed on the transition probability.
Given a sequence of fixed-rate speech features $x_1, x_2, ..., x_T$, our goal is to learn a mapping from a sequence of frames to a sequence of time indices that indicate where the phone boundaries are.

We first formulate an HMM with segment length as transition penalty, utilizing the same duration penalty described in DPDP \cite{kamper2022word, kamper21_interspeech}, and we will refer to this model as HMM-DP.
This HMM consists of $K \times N$ states, where $K$ represents the number of state means (centroids) and $N$ equals the total number of frames $T$.
Each state $s_{k,n}$ represents using the $k^{\text{th}}$ centroid at the $n^{\text{th}}$ segment, which is modeled by a single Gaussian distribution centered at $c_k$ with unit variance.
States sharing the same $k$ also share the same centroid.

We define the emission probability as
\begin{equation}
    P(x_t|z_t = s_{(k,n)}) = \dfrac{1}{(2\pi)^{d/2}} e^{-{\| x_t - c_k \|_2^2}/{2}},
\end{equation}
the state transition probability as
\begin{align}
    P(z_{t} &= s_{(k',n')}|z_{t-1} = s_{(k,n)}) \notag \\
    &\propto \begin{cases}
        e^0          & \text{if } k' = k, n' = n \\
        e^{-\lambda} & \text{if } k' \neq k, n' = n + 1 \\
        e^{-\infty}  & \text{otherwise,}
    \end{cases}
\end{align}
and initial state distribution $P(z_1 = s_{(k,n)})$ which assigns uniform probability for states with $n=1$ as
\begin{align}
    P(z_1 = s_{(k,n)}) \propto \begin{cases}
                    e^{0} & \text{if } n = 1, \; \forall k \\
                    e^{-\infty}  & \text{otherwise}.
                \end{cases}
    \end{align}
In this formulation, transitions are only allowed between $n^{\text{th}}$ and $(n+1)^{\text{th}}$ segments.
Remaining in the same segment requires staying within the same $k$, and incur no additional penalty.
Transitions from the $n^{\text{th}}$ to $(n+1)^{\text{th}}$ segment allow switching $k$, but introduces a penalty parameterized by $\lambda$.
We train this HMM using the Viterbi algorithm with hard decoding (confusingly named segmental k-means \cite{Juang1990segmental}).
The state sequences and the boundaries are identified through backracking, where any frame with a change in $n$ is marked as a boundary.
Although the overall time complexity of this HMM seems to be $O(T \cdot (TK)^2)$, due to the restriction on allowed transitions, the time complexity is reduced to $O(T \cdot TK)$.
As shown in Equation (4), in the first case, there are only $K$ possible transitions from $t-1$ to $t$ for a specific $n$.
For the second case, since the transition probabilities are identical for all pairs of $k$ and $k'$, the weighted forward probabilities to all $k'$ are equal, thus allowing the time complexity to be reduced from $O(K^2)$ to $O(2K)$.

We introduced a second HMM, similar to LBDP, which, unlike DPDP that allows up to $T$ segments, limits the total number of segment to $N$ ($N \leq T$).
We will refer this HMM as HMM-Nseg, indicating the restricted number of segments which also reflected on the reduced number of states of $K \times N$.
This HMM shares the same emission probability, with only a slight variation in the transition probability as
\begin{align}
    P(z_{t} &= s_{(k',n')}|z_{t-1} = s_{(k,n)}) \notag \\
    &\propto 
    \begin{cases}
        e^0        , & \text{if } k' = k, n' = n \\
        e^0        , & \text{if } k' \neq k, n' = n + 1 \\
        e^{-\infty}, & \text{otherwise},
    \end{cases}
\end{align}
where switching $k$ incurs no additional penalty, but increasing $n$ by 1.
With limited number of $N$, it requires the optimization process to identify the most probable transition points with that exact number of segments.
The time complexity of HMM-Nseg could be lower than that of HMM-DP due to the reduced number of states, resulting in $O(T\cdot NK)$.

\vspace{-0.5em}
\subsection{Boundary Features as Transition Penalty}
\vspace{-0.4em}

Building on the success in identifying phone boundaries using Mel spectrograms, we propose incorporating the boundary features from Mel spectrogram into the optimization of the proposed HMM.
Mitchell \textit{et al.} \cite{Mitchell_1995_using} introduced a method for incorporating Spectral Variation Function (SVF) scores into the transition probabilities of HMMs for supervised ASR, utilizing cepstral coefficients for both HMM observations and boundary features.
In contrast, our study explores the intersection of self-supervised features and Mel spectrograms as complementary information sources in HMM training under unsupervised setting.

We first partition the output of SVF, $\tilde{d_1},\tilde{d_2}, \ldots, \tilde{d_T}$, into 0s and 1s by setting a threshold on peak prominence, where a value of 1 indicates a detected boundary.
Using the detected boundaries $y_1, \dots, y_B$, every time frame is assigned the deviation to the closest boundary: $v_t = \min_{b = 1, \dots, B} |t - \hat{y_b}|$.
A linear scaling penalty is then incorporated within the transition probability in the second case of both Equation (2) and (4) when transition of state happens, i.e.,
\begin{equation}
    P_B(z_t|z_{t-1}) \propto P(z_t|z_{t-1}) \times e^{- \gamma \cdot v_t} \quad \text{if } n' = n + 1,
\end{equation}
where $\gamma$ is a hyperparameter adjusting the importance of aligning boundaries in the HMM to those identified from Mel spectrograms, and $P_B$ the state transition probability constrained by boundary features.

\vspace{-0.8em}
\section{Related Work}
\vspace{-0.4em}

We dedicate this section specifically for research on automatic segmentation from the 1970s to the 2000s, as these methods are often overlooked, while many of these ideas remain intriguing and relevant from today's perspective.
While there are three primary approaches for automatic segmentation: peak detection using spectral variation \cite{Wilpon1987investigation,Svendsen1987on,Glass1988multi, falavigna1990dtw, flammia92_icslp}, constrained clustering \cite{glass1988finding, yu2008unsupervised}, and dynamic programming \cite{cohen1981segmenting,Myers1981lbdtw, sharma96_icslp}, we focus on dynamic programming, as it is more closely related to our work.

Dynamic Programming (DP) for unsupervised speech segmentation can be conceptualized as a finite-state machine with constrained transition probabilities, as noted in \cite{cohen1981segmenting}.
One notable approach is the level building dynamic programming (LBDP) \cite{Myers1981lbdtw, sharma96_icslp}.
In \cite{sharma96_icslp}, probability of the observation is modeled using a Gaussian centered at the mean of the frame features within a potential segment.
The LBDP algorithm imposes a constraint on the maximum number of segments allowed in an utterance, and the optimal number of segment is determined using maximum likelihood estimation.

Another interesting approach \cite{cohen1981segmenting} factorizes DP scoring function into the likelihood of a frame being a boundary and the score of segments based on the prior distribution of segment durations.
Boundary scores are calculated with normalized spectral dot product, while segment duration is modeled with a Poisson distribution.
While this approach doesn't explicitly model observation probabilities, it indirectly incorporates them through boundary scores with spectral variation. 

These two DP methods, together with DPDP \cite{kamper2022word}, pose segmentation as an inference problem without relying on any trained parameters.
Building on their success, we proposes an HMM that jointly optimizes the segmentation process with trainable state variables.

\section{Experiments}

We evaluate the proposed HMM with boundary features for unsupervised phone segmentation on TIMIT and Buckeye, both including expert-labeled, time-aligned phone labels.
Following the data processing scripts provided by \cite{kreuk20_interspeech, Strgar2023phoneme}, we use the full training and test set for TIMIT, with 10\% of the training data randomly sampled for validation.
Every utterance in the Buckeye dataset is divided into short segments based on occurrences of VOCNOISE, NOISE, and SIL, resulting in approximately 7.7 hours of processed data \cite{kreuk20_interspeech, Strgar2023phoneme}.
Phone segmentation performance is evaluated using Precision, Recall, F1-score and R-value \cite{rasanen09b_interspeech}, with a boundary tolerance error of 20 ms.
We adopt the strict evaluation protocol described in \cite{cohen1981segmenting,Strgar2023phoneme}, rather than the lenient one commonly used in recent self-supervised methods \cite{kreuk20_interspeech, chorowski21b_interspeech, bhati2022unsupervised}. 
We apply the lenient protocol only when comparing results to previous work.

For both datasets, Mel spectrogram features are extracted using a 25 ms window, a 10 ms stride and 40 Mel filter banks. 
Global mean and variance is calcuated using the respective training set and applied on the Mel spectrogram features.
For self-supervised speech features, we use pretrained HuBERT \cite{hsu2021hubert} and wav2vec 2.0 \cite{BZMA2020}.
Feature from the 9$^{th}$ layer of HuBERT and wav2vec 2.0 are extracted, as it has shown to better correlate to phones \cite{hsu2021hubert, pasad2021layer}.
Since these self-supervised features have a 20 ms stride, we upsample them to match the 10 ms stride of the Mel spectrogram by duplicating each feature in every frame.

\vspace{-0.7em}
\subsection{Self-supervised Features using Peak Detection}

Many recent unsupervised phone segmentation approaches using self-supervised features have utilized peak detection to identify phone boundaries. 
Here, we demonstrate that peak detection may not be the most effective method for self-supervised features when compared to Mel spectrograms. 
A window size of 20 ms is typically used to calculate spectral variation for Mel spectrograms \cite{Glass1988multi, Mitchell_1995_using}.
Instead, we opt for a window size of 30 ms, finding it provides a better indication of phone boundaries, and modified Equation (1) to $d_t = x_{t-2}^\top x_{t+1} / \|x_{t-2}\|\|x_{t+1}\|$.
For both HuBERT and wav2vec 2.0 features, we use a window size of 20 ms, which corresponds to the inherent hop length of the model.

In Table \ref{tab:timit-previous-work}, we compare the performance of Mel spectrograms with the best performing self-supervised models with peak detection. 
The peak prominence threshold is tuned on the respective validation set. 
On both TIMIT and Buckeye, our results show that peak detection with Mel spectrograms significantly outperforms all listed self-supervised models.
This suggests that abrupt acoustic events are more distinctly present at these low-level features. 
Additionally, HuBERT and wav2vec 2.0 comparisons reveal that contrastive learning achieves better performance than mask prediction, explaining the preference for contrastive learning strategies in previous self-supervised methods \cite{kreuk20_interspeech,chorowski21b_interspeech,bhati2022unsupervised,Cuervo2022contrastive}.
Nonetheless, none of these models, regardless of model size or training strategy, can match the performance with Mel spectrograms.

\begin{table}[htb!]
  \caption{Unsupervised phone segmentation using peak detection on \textbf{lenient evaluation}. The models with an asterisk ($^*$) show results reported in the original paper.} 
  \label{tab:timit-previous-work}
  \centering
  \begin{tabular}{llcccc}
    \toprule
    Data & Model & P & R & F1 & RV \\
    \midrule
    \multirow{7}{*}{TIMIT}
    & $^*$CPC \cite{kreuk20_interspeech} & 83.9 & 83.6 & 83.7 & 86.0 \\
    & $^*$ACPC \cite{chorowski21b_interspeech} & 83.7 & 84.7 & 84.7 & 86.9 \\
    & $^*$mACPC \cite{Cuervo2022contrastive} & 84.6 & 84.8 & 84.7 & 86.9 \\
    & $^*$SCPC \cite{bhati2022unsupervised} & 84.6 & \textbf{86.0} & 85.3 & 87.4 \\
    \cmidrule(r){2-6}
    & HuBERT & 66.6 & 66.2 & 66.4 & 71.3 \\
    & wav2vec 2.0 & 68.4 & 74.8 & 71.5 & 74.4 \\
    \cmidrule(r){2-6}
    & log Mel & \textbf{86.9} & \textbf{86.0} & \textbf{86.5} & \textbf{88.4} \\
    \midrule
    \midrule
    \multirow{7}{*}{Buckeye}
    & $^*$CPC \cite{kreuk20_interspeech} & 75.8 & 76.9 & 76.3 & 79.7 \\
    & $^*$ACPC \cite{chorowski21b_interspeech} & 74.7 & 76.6 & 75.6 & 78.9 \\
    & $^*$mACPC \cite{Cuervo2022contrastive} & 74.7 & 76.8 & 75.7 & 79.0 \\
    & $^*$SCPC \cite{bhati2022unsupervised} & 76.5 & \textbf{78.7} & 77.6 & 80.7 \\
    \cmidrule(r){2-6}
    & HuBERT & 62.8  &  65.7  &  64.2  &  68.9 \\
    & wav2vec 2.0 & 64.0  &  69.7  &  66.7  &  70.3 \\
    \cmidrule(r){2-6}
    & log Mel & \textbf{78.6} & \textbf{78.7} & \textbf{78.6} & \textbf{81.8} \\
    \bottomrule
  \end{tabular}
  \vspace{-0.5em}
\end{table}

\vspace{-0.7em}
\subsection{Proposed HMMs}

For all proposed HMMs, we use $K=50$ for all experiments.
These models are trained for 10 epochs on TIMIT and 20 epochs on Buckeye.
We denote the HMM with boundary features (BF) as a transition penalty as HMM-Nseg-BF and HMM-DP-BF for future reference.
In the HMM-Nseg and HMM-Nseg-BF approaches, given the variable length of utterances, we avoid setting the same fixed number of segments $N$ for all utterances.
Instead, the number of segments is determined by an average duration $L$, allowing the number of segments to be calculated dynamically.
The hyperparameters, average phone duration $L$, duration penalty $\lambda$, and boundary features $\gamma$, are tuned using the validation set.\footnote{For TIMIT, we set $L=8.1$ for HMM-Nseg, $\lambda=1.9$ for HMM-DP, $L=8.1, \gamma=1.2$ for HMM-Nseg-BF, $\lambda=0.4, \gamma=0.9$ for HMM-DP-BF using HuBERT. For Buckeye, we use $L=8.5$ for HMM-Nseg, $\lambda=2.2$ for HMM-DP, $L=8.1, \gamma=1.0$ for HMM-Nseg-BF and $\lambda=0.5, \gamma=1.0$ for HMM--DP-BF with HuBERT.}

Our initial analysis compares the performance of peak detection with HMM-DP and HMM-Nseg using HuBERT and wav2vec 2.0 (W2V2) features to determine whether the HMM-based system is a better fit for these features. 
The results are shown in Table \ref{tab:hmm-scores}.
Starting with HuBERT, both HMM-DP and HMM-Nseg significantly outperform peak detection, showing R-value (RV) improvements of 10\% on TIMIT and 9\% on Buckeye absolute. 
This suggests that the underlying phone structure in the HuBERT feature space may be well represented by a single Gaussian. 
Conversely, W2V2 features show slightly worse performance when using HMM, indicating that the contrastive nature of these features might not cluster phones based on Euclidean distance, making a single Gaussian model less effective.

\begin{figure}[htb!]
    \centering
    \includegraphics[width=\linewidth]{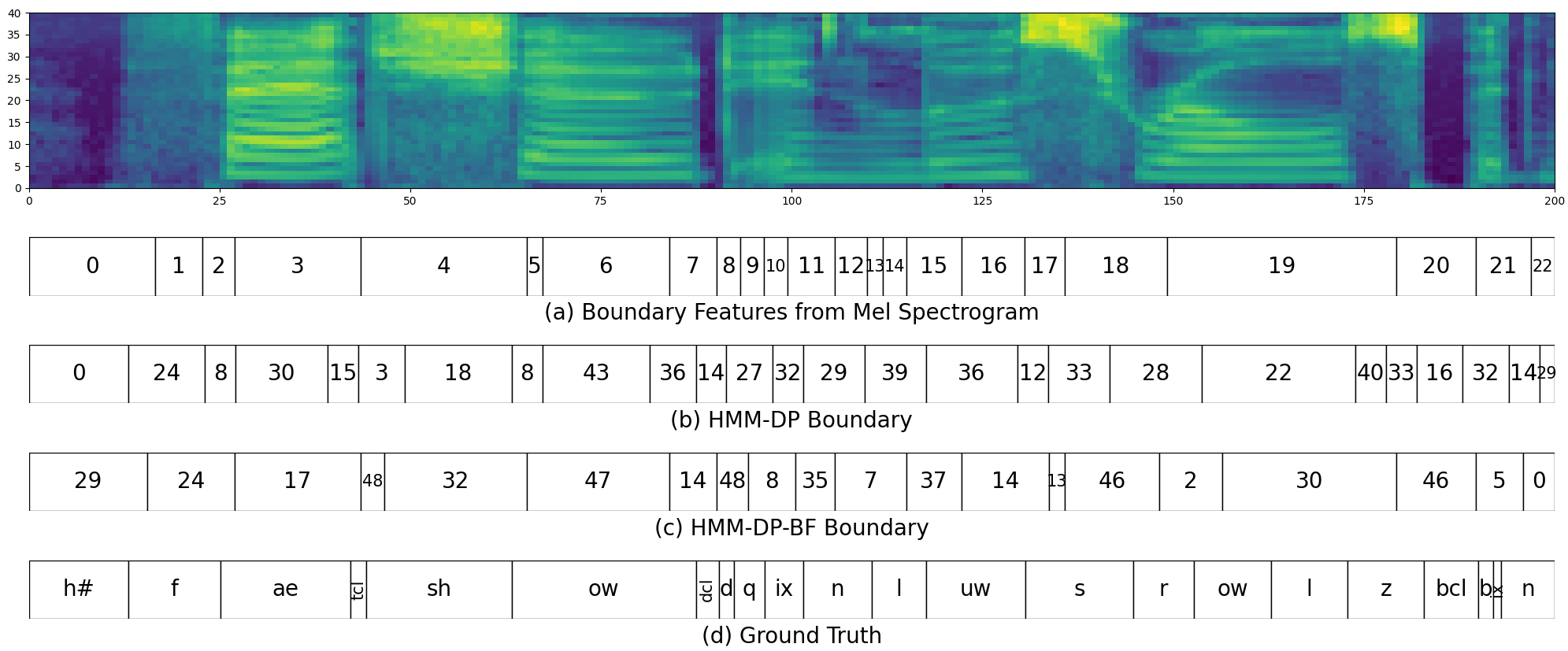}
    \caption{Comparison of the detected boundaries by different HMMs using HuBERT features on fadg0\_si1909 from TIMIT.}
    \label{fig:hmm-bf}
    \vspace{-1.5em}
\end{figure}

Next, we evaluate the impact of incorporating boundary features from Mel spectrograms with self-supervised features on HMM. 
The results demonstrate significant improvements for both HuBERT and W2V2, with HMM-DP-BF and HMM-Nseg-BF outperforming both peak detection and HMM without boundary features.
An example of the resulting boundary refinement using boundary features is shown in Fig. \ref{fig:hmm-bf}. 
We observe a notable difference between boundaries detected from Mel spectrograms and those from HMM-DP using HuBERT.
By integrating both features into the HMM, the resulting segmentation more closely aligns with the ground truth boundaries.
This results in improvements in both precision and recall, leading to a 6\% absolute improvement in the R-value for HuBERT and a 12\% absolute improvement for W2V2 on both TIMIT and Buckeye.

We compared our HMMs against the neural network method proposed in \cite{Strgar2023phoneme}, which uses noisy boundary labels derived from a previous self-supervised model \cite{kreuk20_interspeech} as targets and applies frame-wise binary cross-entropy (BCE) as its learning objective.
The use of noisy boundary labels is conceptually similar to our use of boundary features.
The neural network approaches either fine-tune all 12 layers of the pre-trained model, or use a 5-layer CNN combined with layer-specific CNNs applied to layer-wise features, totaling 65M parameters.
As the authors suggested, the readout mode with 5-layer CNN performs better than fine-tuning, and the results reported by the authors are listed in Table \ref{tab:hmm-scores}.
Our HMMs perform on par with their best-performing methods, with the advantage of requiring only $50 \times 768$ parameters and a much faster runtime in both training and decoding.

\begin{table}[htb!]
  \caption{Unsupervised phone segmentation using HMMs with \textbf{strict evaluation}. The model with an asterisk ($^*$) show results reported in the original paper.}
  \label{tab:hmm-scores}
  \centering
  \begin{tabular}{llcccc}
    \toprule
    Model & Alg. & P & R & F1 & RV \\
    \midrule
    \multicolumn{6}{c}{\textbf{TIMIT}} \\
    \midrule
    log Mel & Peak & 80.4 & 79.3 & 79.8 & 82.8 \\  
    \midrule
    HuBERT & Peak & 63.9 & 60.8 & 62.3 & 68.1 \\
    HuBERT & HMM-NSeg & 76.0 & 75.2 & 75.6 & 79.2 \\
    HuBERT & HMM-DP & 73.7 & 77.4 & 75.5 & 78.7 \\
    HuBERT & HMM-NSeg-BF & \textbf{84.9} & 78.3 & 81.4 & 83.5 \\
    HuBERT & HMM-DP-BF & 84.1 & \textbf{80.1} & \textbf{82.1} & \textbf{84.4} \\
    \midrule
    W2V2 & Peak & 67.1 & 67.8 & 67.4 & 72.1 \\
    W2V2 & HMM-NSeg & 66.8 & 66.1 & 66.4 & 71.4 \\
    W2V2 & HMM-DP & 65.5 & 69.1 & 67.3 & 71.5 \\
    W2V2 & HMM-NSeg-BF & 82.9 & 78.6 & 80.7 & 83.2 \\
    W2V2 & HMM-DP-BF & \textbf{83.3} &  \textbf{79.6} &  \textbf{81.4} &  \textbf{83.9}  \\
    \midrule
    $^*$HuBERT & Frame BCE \cite{Strgar2023phoneme} & 82.4 &  \textbf{81.2} &  \textbf{81.8} &  \textbf{84.5} \\
    $^*$W2V2 & Frame BCE \cite{Strgar2023phoneme} &  \textbf{84.9} & 78.5 & 81.6 & 83.7 \\
    \midrule
    \midrule
    \multicolumn{6}{c}{\textbf{Buckeye}} \\
    \midrule
    log Mel & Peak       & 74.1 & 75.0 & 74.6 & 78.2 \\  
    \midrule
    HuBERT & Peak        & 61.6 & 60.3 & 61.0 & 66.8 \\
    HuBERT & HMM-NSeg    & 70.4 & 71.1 & 70.8 & 75.0  \\
    HuBERT & HMM-DP      & 70.7 & 71.6 & 71.2 & 75.3  \\
    HuBERT & HMM-NSeg-BF & 78.6 & \textbf{76.4} & 77.5 & 80.8 \\
    HuBERT & HMM-DP-BF   & \textbf{81.0} & 75.5 &  \textbf{78.1} &  \textbf{81.0}  \\
    \midrule
    W2V2 & Peak          & 62.8 & 64.9 & 63.8 & 68.8 \\
    W2V2 & HMM-NSeg      & 60.4 & 60.9 & 60.7 & 66.3  \\
    W2V2 & HMM-DP        & 60.1 & 62.3 & 61.2 & 66.5  \\
    W2V2 & HMM-NSeg-BF   & \textbf{77.7} & \textbf{72.5} &  \textbf{75.0} &  \textbf{78.5} \\
    W2V2 & HMM-DP-BF     & 76.7 & 72.1 & 74.3 & 78.0  \\
    \midrule
    $^*$HuBERT & Frame BCE \cite{Strgar2023phoneme} & 75.3 &  \textbf{79.4} & 77.3 & 80.1 \\
    $^*$W2V2 & Frame BCE \cite{Strgar2023phoneme} & \textbf{77.9} & 77.4 &  \textbf{77.7} &  \textbf{81.0} \\
    \bottomrule
  \end{tabular}
  \vspace{-1em}
\end{table}

\subsection{HMM Training vs. Two-stage Decoding}

A major distinction between our method and previous work (both LBDP \cite{sharma96_icslp} and DPDP \cite{kamper2022word}) is that in our HMM, the parameters (the centroids) are jointly learned with the constrained transition probability. 
In contrast, previous methods use a pre-defined set of quantized vectors for the centroids, often derived from pre-clustered k-means or a VQ codebook trained during self-supervised learning \cite{kamper2022word}.
In essence, LBDP and DPDP approach segmentation as an inference problem, whereas we treat it as a learning problem.

To better compare our method with both LBDP and DPDP, we train an offline k-means clustering with $k=50$ and use the learned centroids in our HMM for inference. 
We treat these methods as a two-stage decoding process, with the first stage being the k-means clustering step and the second stage being the HMM inference.
We rename DPDP to VQ-DP, to emphasize the inference using quantized vectors with a duration penalty. 
For other variations using k-means centroids, we use the names VQ-Nseg, VQ-Nseg-BF, and VQ-DP-BF, where VQ-Nseg is equivalent to LBDP.

Table \ref{tab:vq-all} presents the two-stage decoding (VQ) results.
Comparing our proposed HMMs to the VQ approaches, we observe consistent improvements with the HMMs. 
Without boundary features, HMM with HuBERT results in a 4-6\% absolute improvement in R-value compared to VQ, while no significant improvements are observed with W2V2.
Among these VQ methods, HuBERT features consistently outperform W2V2, which again suggests that the W2V2 feature space may not be well-modeled with a simple Gaussian distribution.
Additionally, boundary features (BF) prove beneficial even just for inference, showing an improvement of around 10\% absolute in F1 and R-value, particularly for W2V2 features.

\begin{table}[t!]
  \caption{Unsupervised phone segmentation using two-stage decoding (VQ) with \textbf{strict evaluation}.}
  \label{tab:vq-all}
  \centering
  \begin{tabular}{llcccc}
    \toprule
    Feat. & Alg. & P & R & F1 & RV \\
    \midrule
    \multicolumn{6}{c}{\textbf{TIMIT}} \\
    \midrule
    HuBERT & VQ-NSeg & 67.9 & 68.9 & 68.4 & 73.0 \\
    HuBERT & VQ-DP & 68.1 & 70.5 & 69.3 & 73.5 \\
    HuBERT & VQ-NSeg-BF & 86.0 & 76.4 & 80.9 & 82.5 \\
    HuBERT & VQ-DP-BF & 85.9 & 77.0 & 81.2 & 82.9  \\
    \midrule
    W2V2 & VQ-NSeg & 65.4 & 66.3 & 65.8 & 70.7 \\
    W2V2 & VQ-DP & 65.7 & 66.6 & 66.2 & 71.0 \\
    W2V2 & VQ-NSeg-BF & 85.1 & 76.4 & 80.5 & 82.4 \\
    W2V2 & VQ-DP-BF & 84.8 & 77.1 & 80.7 & 82.8 \\
    \midrule
    \midrule
    \multicolumn{6}{c}{\textbf{Buckeye}} \\
    \midrule
    HuBERT & VQ-NSeg & 64.3 & 67.3 & 65.8 & 70.3  \\
    HuBERT & VQ-DP & 65.6 & 66.9 & 66.2 & 71.0 \\
    HuBERT & VQ-NSeg-BF & 78.9 & 72.3 & 75.4 & 78.7 \\
    HuBERT & VQ-DP-BF & 78.3 & 73.4 & 75.7 & 79.1  \\
    \midrule
    W2V2 & VQ-NSeg & 59.8 & 61.7 & 60.7 & 66.1 \\
    W2V2 & VQ-DP & 59.9 & 61.8 & 60.8 & 66.2 \\
    W2V2 & VQ-NSeg-BF & 78.1 & 71.8 & 74.8 & 78.2 \\
    W2V2 & VQ-DP-BF & 77.4 & 73.2 & 75.2 & 78.8 \\
    \bottomrule
  \end{tabular}
  \vspace{-1em}
\end{table}

\subsection{HMM Phone Purity Analysis}

Our proposed HMMs, designed for unsupervised phone segmentation, also play a significant role in acoustic unit discovery \cite{varadarajan2008unsupervised,lee2012nonparametric,kamper2017segmental,kamper2017embedded}.
The cluster assignment $k$ for each frame can be interpreted as the discovered acoustic units \cite{hsu2021hubert,wells2022phonetic}, and we aim to explore its correlation with phone labels \cite{Yang2023towards}.
To assess this, we measure frame-wise phone purity and cluster purity, examining the degree to which the state assignments align with phone labels following \cite{hsu2021hubert}.
Phone purity reflects the overall accuracy where frames are assigned to phone labels based on their clusters, and each cluster's phone label is determined by the majority phone in that cluster. 
This metric shows the upper bound of frame-wise accuracy if assigning a single phone label to each cluster.
Cluster purity, on the other hand, increases when the frames of a single phone predominantly reside within one cluster.
We will concentrate primarily on phone purity, as it reflects the phone error rate when each cluster is treated as a distinct phone.
For the TIMIT dataset, we use the original set of 61 phones, and for the Buckeye dataset, we evaluate using the original set of 75 phones, including noise and silence labels.

We first evaluate purity metrics by comparing models without boundary features, i.e., k-means clustering, two-stage decoding (VQ), and HMMs, as shown in the top 3 rows of each block in Table \ref{tab:purity}.
Although the centroids of VQ are identical to those from k-means clustering, the segment constraint brings a consistent improvement in both phone purity and cluster purity across all configurations.
Moreover, the HMM approaches significantly outperform both k-means and VQ, achieving a 4\% absolute improvement in phone purity with HuBERT and 2\% absolute with W2V2 in both datasets.

Additionally, incorporating boundary features in HMMs further improves phone purity.
While the improvement on the TIMIT dataset is modest, phone purity on the Buckeye dataset increases by 2-3\% absolute.
This suggests that boundary features not only improve segmentation, but also improve the alignment of phone labels with their respective clusters.
Given that HuBERT codes from k-means clustering are widely used as speech tokens in various tasks, our findings suggest that HMM states provide even better alignment to phones. 
This, along with the improved segmentation, highlights the potential of HMMs to significantly assist in the understanding of speech and its nuanced phonetic structure.

\begin{table}[tbp!]
    \caption{Phone Purity (PP) and Cluster Purity (CP) evaluated using different segmentation algorithms with HuBERT and W2V2 on the TIMIT and Buckeye datasets.}
    \label{tab:purity}
    \centering
    \begin{tabular}{llcccc}
        \toprule
        \textbf{TIMIT} & & \multicolumn{2}{c}{VQ} &  \multicolumn{2}{c}{HMM}  \\
        \cmidrule(r){3-4} \cmidrule(r){5-6}
        Feat & Alg. & PP & CP & PP & CP \\
        \cmidrule(r){1-6}
        HuBERT & K-means & 47.3 & 42.1 & - & -  \\
        HuBERT & Nseg & 47.5 & 43.3 & 51.6 & 49.9 \\
        HuBERT & DP & 47.7 & 43.6 & 51.6 & 48.6 \\
        HuBERT & Nseg-BF & 47.9 & 48.1 & \textbf{52.1} & \textbf{50.2} \\
        HuBERT & DP-BF & 48.0 & 48.2 & 51.5 & 49.2 \\

        \cmidrule(r){1-6}
        W2V2 & K-means & 43.3 & 39.0 & - & - \\
        W2V2 & Nseg & 44.2 & 41.1 & 45.6 & 39.7 \\
        W2V2 & DP & 44.3 & 41.3 & 46.6 & 41.5 \\
        W2V2 & Nseg-BF & 45.4 & 44.0 & \textbf{48.9} & 42.4 \\
        W2V2 & DP-BF & 45.4 & 44.0 & 47.7 & \textbf{43.2} \\
        
        \midrule
        \midrule
        
        \textbf{Buckeye} & & \multicolumn{2}{c}{VQ} &  \multicolumn{2}{c}{HMM}  \\
        \cmidrule(r){3-4} \cmidrule(r){5-6}
        Feat & Alg. & PP & CP & PP & CP \\
        \cmidrule(r){1-6}
        HuBERT & K-means & 42.2 & 34.2 & - & -  \\
        HuBERT & Nseg & 42.4 & 36.4 & 46.6 & \textbf{42.8} \\
        HuBERT & DP & 42.4 & 36.6 & 45.7 & 42.3 \\
        HuBERT & Nseg-BF & 42.9 & 37.9 & \textbf{49.4} & 41.8 \\
        HuBERT & DP-BF & 42.9 & 38.0 & 48.1 & 40.2 \\

        \cmidrule(r){1-6}
        W2V2 & K-means & 35.6 & 29.1 & - & -  \\
        W2V2 & Nseg & 36.0 & 30.9 & 38.3 & 32.3 \\
        W2V2 & DP & 36.1 & 31.0 & 38.0 & 33.2 \\
        W2V2 & Nseg-BF & 37.3 & 35.1 & \textbf{40.7} & \textbf{33.8} \\
        W2V2 & DP-BF & 37.4 & 35.1 & 40.6 & 32.5 \\
        
        \bottomrule
    \end{tabular}
    \vspace{-1em}
\end{table}

\section{Conclusion}

We propose a simple HMM for unsupervised phone segmentation and show its strong performance compared to approaches that rely on training neural networks of several layers. 
Our HMM not only excels in unsupervised phone segmentation but also shows improved phone purity in the discovered units. 
Our results suggest that past wisdom in unsupervised phone segmentation should not be neglected, and simple approaches might be just as good if not better than deep learning approaches that we are too accustomed to now.

\clearpage

\bibliographystyle{IEEEbib}
\bibliography{ref}

\end{document}